\documentclass[10pt,twocolumn,letterpaper]{article}

\usepackage{lib/cvpr}
\usepackage{times}
\usepackage{epsfig}
\usepackage{graphicx}
\usepackage{amsmath}
\usepackage{amssymb}
\usepackage{booktabs}
\usepackage{amsmath}
\usepackage{multirow}

\usepackage[pagebackref=true,breaklinks=true,colorlinks,bookmarks=false]{hyperref}

\cvprfinalcopy 

\def\cvprPaperID{19} 

\ifcvprfinal\fi

\begin{document}
\title{
    Exploring Racial Bias within Face Recognition via per-subject Adversarially-Enabled Data Augmentation
}

\author{
    Seyma Yucer\textsuperscript{1},
    Samet Ak\c{c}ay\textsuperscript{1,3},
    Noura Al-Moubayed\textsuperscript{1},
    Toby P. Breckon\textsuperscript{1,2} \\    
    Department of \{Computer Science\textsuperscript{1},
    Engineering\textsuperscript{2}\}, Durham University, Durham, UK\\
    COSMONiO\textsuperscript{3}, Durham, UK\\
    {
    \tt\small
    \{
        \href{mailto:seyma.yucer-tektas@durham.ac.uk}
             {seyma.yucer-tektas},        
        \href{mailto:samet.akcay@durham.ac.uk}
             {samet.akcay},
        \href{mailto:noura.al-moubayed@durham.ac.uk}
             {noura.al-moubayed},
        \href{mailto:toby.breckon@durham.ac.uk}
             {toby.breckon}
    \}@durham.ac.uk
    }  
}

\maketitle

\begin{abstract}
    Whilst face recognition applications are becoming increasingly prevalent within our daily lives, leading approaches in the field still suffer from performance bias to the detriment of some racial profiles within society. In this study, we propose a novel adversarial derived data augmentation methodology that aims to enable dataset balance at a per-subject level via the use of image-to-image transformation for the transfer of sensitive racial characteristic facial features. Our aim is to automatically construct a synthesised dataset by transforming facial images across varying racial domains, while still preserving identity-related features, such that racially dependant features subsequently become irrelevant within the determination of subject identity. We construct our experiments on three significant face recognition variants: Softmax, CosFace and ArcFace loss over a common convolutional neural network backbone. In a side-by-side comparison, we show the positive impact our proposed technique can have on the recognition performance for (racial) minority groups within an originally imbalanced training dataset by reducing the per-race variance in performance.
\end{abstract}
\section{Introduction}
\label{sec:introduction}

Numerous machine learning applications utilising facial attributes have proliferated in recent years as autonomous decision-making processes have become widely adopted by companies and governments \cite{masi2018deep}. A growing number of applications based on face analyses for surveillance \cite{bashbaghi2019deep}, recruitment \cite{hemamou2019hirenet}, and health-care \cite{uddin2020depression} have increasingly become integrated into our daily lives.

However, the generalisation of such research and applications is problematic due to the prevalence of bias occurrences within face recognition \cite{raji2019actionable}. The imbalance in specific demographic groups occurring with varying geographic locale globally, including race, age or gender, poses a challenge of transparent explanations and solutions for facial recognition applications. Hence, to cope with real-world diversity, it is crucial to have a profound understanding of this bias within every aspect \cite{buolamwini2018gender}.
\begin{figure}
\label{fig:introduction}
\includegraphics[scale=0.33]{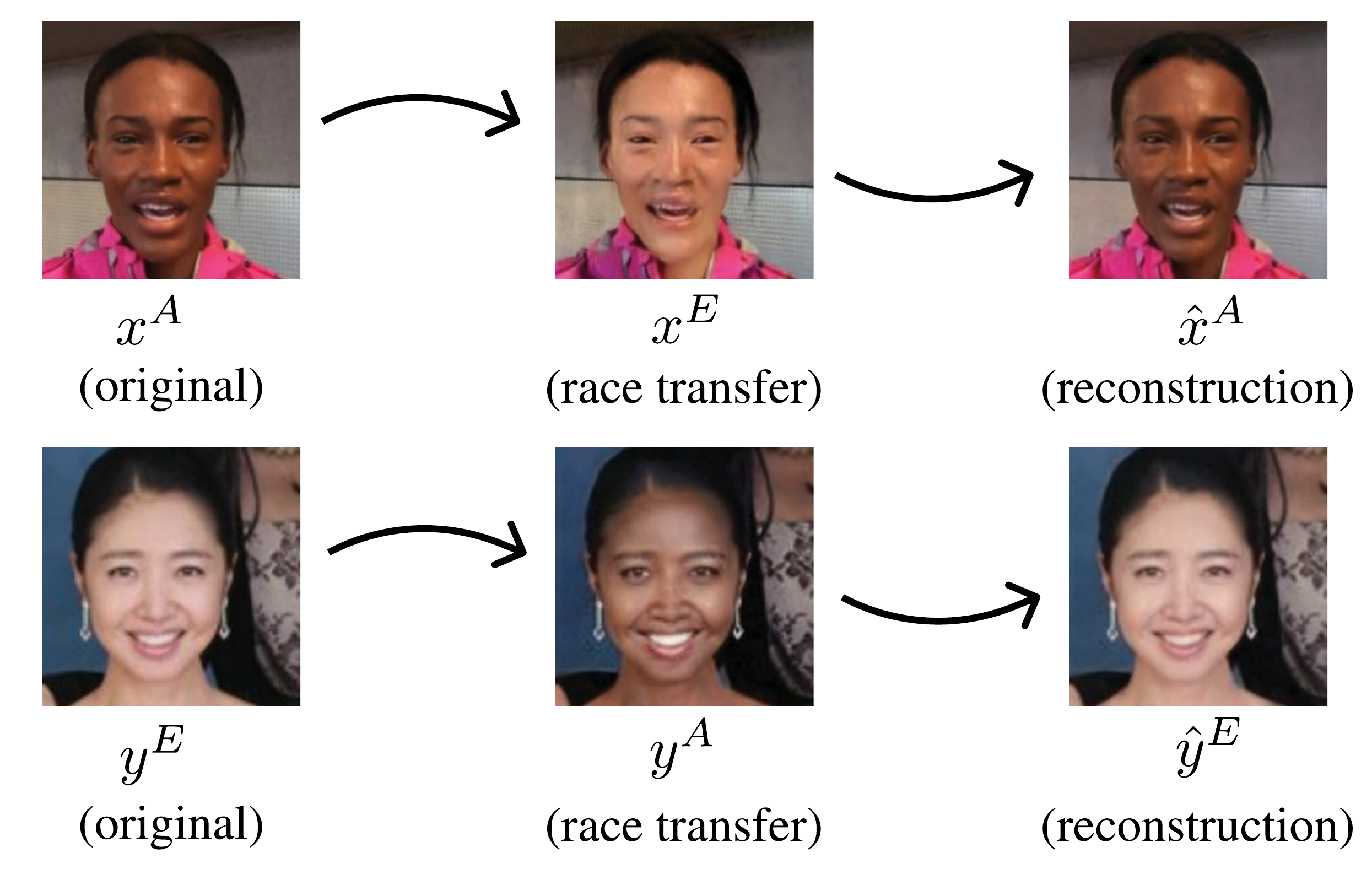}
\caption{ Racial transformation example using \cite{zhu2017unpaired}. We transfer an African image $x^A$ to Asian image $y^E$ and obtain sythesised $x^E$ in Asian domain and we reconstruct $\hat{x}^A$ from $x^E$ image. Asian image $y^E$ to African image $x^A$ transformation follows the same procedure.
} \end{figure} 

Bias in machine learning has been extensively studied for decades \cite{pedreshi2008discrimination,barocas2017fairness}. These studies provide the fundamental understanding of the underlying reasons for face recognition bias which has also seen a surge of interest in recent years \cite{garcia2019harms,buolamwini2018gender}. Studies have addressed this problem in various perspectives such as data pre-processing \cite{yang2020towards,ali2019mfc,sadhukhan2019learning}, and adversarial training \cite{kim2019learning,gong2019debface,wang2019approaching}.

Meanwhile, recent advances in Generative Adversarial Networks (GAN), have led to realistic image generation \cite{karras2019style} and even class generation \cite{ali2019mfc}. Such advances in the field have a promising potential to overcome the bias in face recognition via realistic image generation as most of the face recognition datasets have a significantly imbalance distribution on either classes \cite{huang2019deep} or demographic groups \cite{hupont2019demogpairs}.

In this study, we address the racial bias of face recognition from an adversarial augmentation point of view. As most of the datasets \cite{wang2019racial,wang2019skewness,buolamwini2018gender} consist of four major racial groups, namely African, Asian, Caucasian and Indian, we seek group-fairness among these races, in terms of facial recognition performance, by utilising generative adversarial network (GAN) \cite{goodfellow2014generative}.

Previous work \cite{kim2019learning,gong2019debface,wang2019approaching} has established adversarial techniques to minimise mutual information on identity features, which reveal sensitive attributes about race, gender and age of the subject. However, such approaches \cite{kim2019learning,mcduff2019characterizing}, have failed to effectively address the trade-off between suppressing the use of such sensitive attributes and the loss of key identity-related features which pertain to the overall performance of the facial recognition approach. Our solution, instead, uses an adversarial image re-synthesise technique \cite{zhu2017unpaired}, to transform sensitive attributes across a set of synthetic images comprising the full range of races being considered within the facial recognition problem. 

By doing so, we preserve the important identity-related features whilst making the racially dependent features of the face less prevalent due to the artificially synthesised distribution of these identity characteristics across the full range of race profiles for any given individual.

Figure \ref{fig:introduction} illustrates how we transform the identity characteristics, and hence features, any given individual across multiple racial profiles using a CycleGAN \cite{zhu2017unpaired}. It proposes transformation across racial domains and reconstruction to produce an identical image from a transformed image during the cyclic adversarial training. 

To show its robustness, we explore the performance of our approach using balanced and imbalanced training datasets. The main contributions of this paper are as follows:

\begin{itemize}

    \item we propose an adversarial image-to-image transformation technique to mitigate racial bias based on the cyclic adversarial training approach of  CycleGAN \cite{zhu2017unpaired}.
    \vspace{-0.50cm}
    \item we illustrate both quantitative and qualitative performance of our proposed facial data augmentation techniques over established benchmark datasets within the face recognition domain, establishing a statistical paradigm for the presentation of recognition results on a per-race basis.
\end{itemize}

The rest of this paper is structured as follows: in Section \ref{sec:related-work}, we review the current solutions for face recognition bias in three different categories.
We present a methodology for this study in Section \ref{sec:proposedmethod} with our experimental setup and results in Section \ref{sec:experimental-setup} and \ref{sec:results}, respectively. An extended discussion on adversarial face recognition bias for both balanced and imbalanced datasets is presented within Section \ref{sec:results} with our final conclusions subsequently presented in Section \ref{sec:conclusion}.

\section{Related Work}
\label{sec:related-work}
Bias and fairness in machine learning have been studied in the last decade, and significant research \cite{suresh2019framework,mehrabi2019survey} draws attention to bias for different fields like face recognition, action recognition or language processing. 

As one of the most prominent fields of machine learning, face recognition has been extensively used across different areas \cite{wang2017face,stephen2017facial}. As the popularity of face recognition increases, we face more bias incidents \cite{garcia2019harms}. Moreover, studies \cite{buolamwini2018gender,nagpal2019deep,cavazos2019accuracy} point out the bias of current face recognition web services and state-of-the-art algorithms for demographic groups such as age, gender, and race. Although definitions of demographic attributes might be uncertain, it is still important to strive for group-fairness \cite{zemel2013learning}.

Studies of bias in face recognition which use contemporary deep learning approaches are categorised into three main groups: pre-processing (data preparation), in-processing (model training) and post-processing (output inference) techniques.

\noindent \textbf{Pre-processing Methods.} Previous studies \cite{merler2019diversity,hupont2019demogpairs} revealed that the public face recognition datasets have more male and lighter skin tone subjects than respectively female and darker skin tone subjects. This is due to the images within these datasets are mostly from celebrities, including sports players, actors, politicians, collected from predominantly white male subjects. In other respects, the studies of \cite{wang2019racial,merler2019diversity} released balanced datasets for four racial groups; they do not provide universal race coverage nor they are not openly and readily available for access.

To obtain fair datasets, studies \cite{kamiran2010classification,ali2019mfc,sadhukhan2019learning} propose re-sampling methods by either dropping or augmenting samples in the datasets. Downsampling can be considered as a solution for avoiding bias despite the information loss it introduces. Augmentation techniques \cite{ali2019mfc,sadhukhan2019learning} for image generation have improved significantly using adversarial learning. However, the limitations, as described in \cite{jain2020imperfect}, are still a concern for mitigating bias. Feature transformation is another pre-processing  approach \cite{yin2019feature} that improves the feature space of under-represented subjects by moving the distribution of the feature space closer to the regular, supposedly unbiased distribution. 

\noindent \textbf{In-processing Methods.} In-processing methods are divided into three groups: (i) adversarial approaches \cite{kim2019learning,mcduff2019characterizing,gong2019debface,wang2019approaching}, (ii) domain adaptation methods \cite{wang2019racial} and (iii) cost-sensitive learning techniques \cite{khan2017cost,baloch2019focused}. Adversarial methods focus on sensitive features on the image; with \cite{kim2019learning} proposing an adversarial feature learning approach rather than learning all the feature representations from the image. In this way, it minimises mutual information between bias features and characteristic features to decrease bias influence. The experiments of \cite{kim2019learning} are relatively simplistic compared to face recognition bias. Distinguishing demographic information within an image is a serious trade-off of face recognition as demographic features (age, gender, race), and identity features overlap. Another approach in \cite{mcduff2019characterizing}, addresses this problem by highlighting the difficulty of setting a  demographic condition in realistic face generation.
On the other hand, \cite{gong2019debface} debiases images by minimising correlation on disentangled features. Another study \cite{wang2019approaching} reduces the dependence on sensitive attributes. Despite achieving state-of-the-art results on the test, there is still ample room for further understanding of bias.

A domain adaptation technique, \cite{wang2019racial}, transfers the Caucasian domain to non-Caucasian domains during training but requires to have at least one source domain to transfer into others. Cost-sensitive solutions \cite{khan2017cost,baloch2019focused} have been used for imbalanced learning and machine learning fairness in general. For face recognition, adaptive margin \cite{deng2019arcface} or cluster large margin settings \cite{huang2019deep} are more frequently considerable since the aim is to have intra-class compactness and inter-class discrepancy for large scale datasets. Distinguishing the group features on hypersphere helps to avoid overfitting of under-represented groups. Adaptive margins \cite{wang2019skewness} for each race improves the scatter of features of races.

\noindent \textbf{Post-processing Methods.} Post-processing studies are based on either detecting the bias or improving the fairness after training the model. For example, \cite{kim2019multiaccuracy} proposes a Multiaccuracy-Boost algorithm for any machine learning algorithms to improve fairness. IBM \cite{bellamy2018ai} provides an extensive toolkit to detect bias and determine the current model fairness level. For broader explanations, \cite{srinivas2019face,cavazos2019accuracy} give demographic bias level of current state-of-the-art face recognition algorithms.

Motivated by \cite{zhu2017unpaired}, our approach is based on adversarial image synthesise to mitigate bias. Unlike other adversarial studies \cite{kim2019learning,wang2019approaching}, we transform race information from one group to another for fair face recognition. We aim to augment sensitive attributes to make them irrelevant for face recognition solutions.

    \section{Proposed Method}
\label{sec:proposedmethod}

We present our methodology in three parts: we first describe our problem definition in Section \ref{subsec:problem_definition}, explain image-to-image transfer method \cite{zhu2017unpaired} for race transformation to mitigate face recognition bias in Section \ref{subsec:cyclegan} and outline our comparator state-of-the-art face recognition algorithms \cite{wang2018cosface,deng2019arcface} in Section \ref{subsec:facerecognition}.

\begin{figure*}[tbp]
\center

\includegraphics[width=\textwidth]{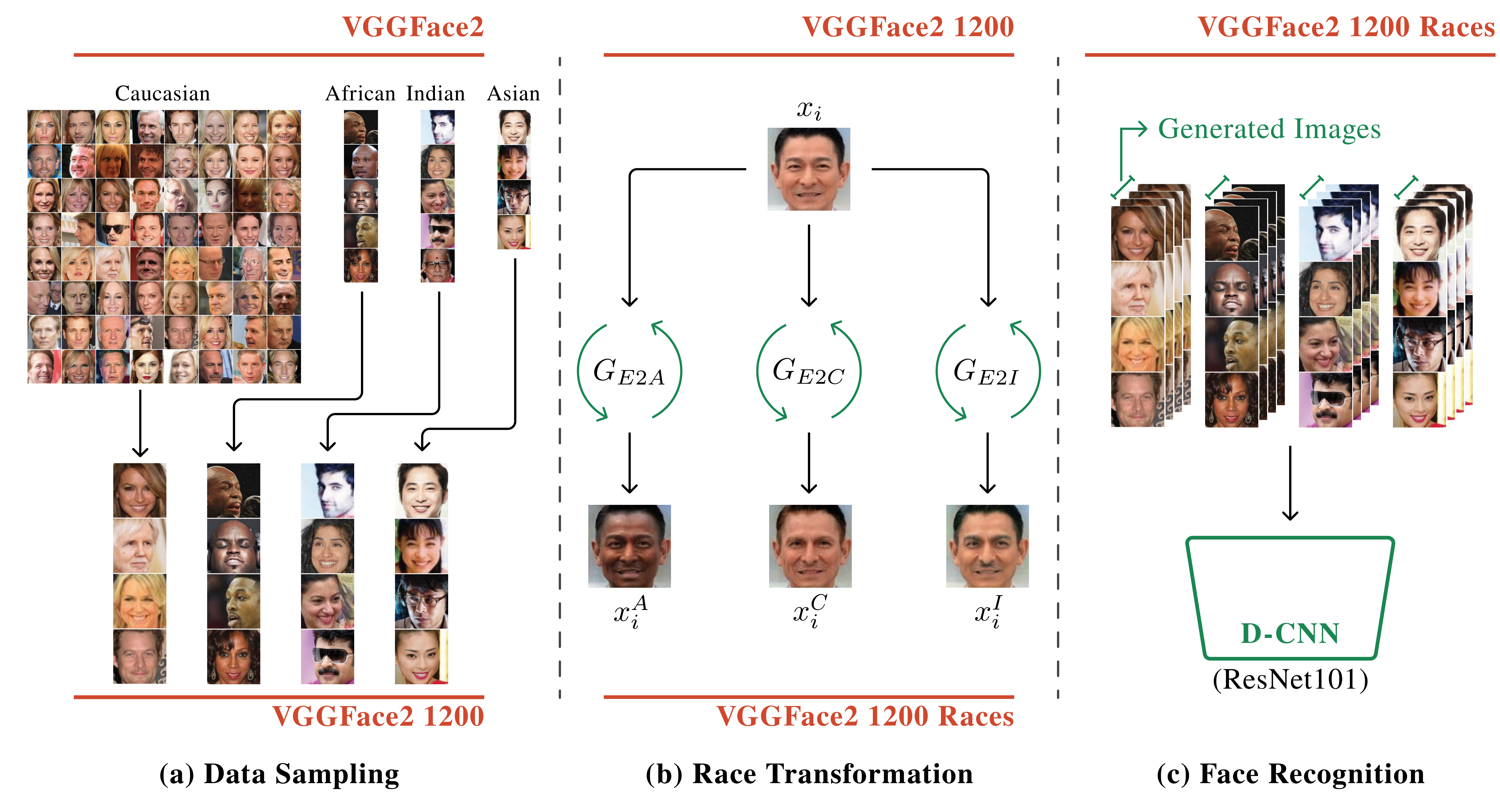}
\caption{Overview of our solution in three phases: (a) describes imbalanced distribution of VGGFace2 \cite{cao2018vggface2} and downsampling it to VGGFace2 1200. (b) illustrates race domain transformation schema for a given image $x_i$ (c) shows face recognition algorithms with Softmax \cite{liu2017sphereface}, CosFace \cite{wang2018cosface} and ArcFace \cite{deng2019arcface} loss functions using VGGFace2 1200 Races.} 
\label{fig:overview}
\end{figure*}

\subsection{Problem Definition}
\label{subsec:problem_definition}
In this section, we define our problem by introducing the general terms of machine learning bias. \textit{Disparate impact}, as indirect discrimination, appears when there is a correlation between sensitive attributes (age, gender, race) and other attributes. It causes inequality on outcomes for different demographic groups, as observed on various machine learning applications, including face recognition web services \cite{buolamwini2018gender}.

Ideally, a machine learning algorithm should require that the conditional probability $P$ of the output given input $x$ does not depend on any \textit{sensitive attributes} which is demographic features in our case. This \textit{unawareness} can be formalized as $P(y \mid x) = P(y \mid x,s)$, where $x$ is an input, $y$ is the corresponding label and $s$ is a sensitive attribute that does not alter the outcome. However, removing dependency is highly challenging for face recognition due to high \textit{mutual information} between facial features and sensitive attributes, like race.

For a given face image dataset, $D= \left[ x_1,x_2,x_3, \dotsc, x_N \right]$ provides $N$ number of face images. A feature embedding vector of an image, $z_i = \left[ f_1,f_2, \dotsc,f_d \right]$, where $z_i \in \mathbb{R}^d$, is commonly statistically dependent on sensitive attributes where it causes \textit{indirect discrimination} for particular demographic groups which potentially form overlapping, subsets of $D$. Although the common approach for face recognition bias is to minimise this mutual information to remove the dependency on sensitive features; it is still an extremely difficult task using face features without sacrificing any prior information for face recognition as shown in \cite{kim2019learning, mcduff2019characterizing}. 

Hence, we approach the problem from a completely different perspective by transferring sensitive attributes from one domain to another whilst simultaneously preserving prior information for recognition. On the other hand, we are aware that some features are more prevalent in some demographic groups than others. The sensitive information, in this case, may improve the prior information for the recognition task. Lighter skin allows the model to learn more detailed features given characteristics of modern cameras and common scene lumination conditions. A novel input mechanism which projects different sensitive information for one image to a model makes race modelling irrelevant. As a result, we ask a question;\textit{ What if we augment and transfer sensitive information rather than removing it$?$}  To answer this question, we present a new pre-processing based method requires augmentation of sensitive attributes of an image.

Our new inputs consist of three generated images from different domains for each image. Given the race domains $\{A, E, C, I\}$ for $\{African, Asian, Caucasian, Indian\}$ respectively, we aim to transform an image $x_i$ from one domain as an image $x_j$ to another domain. For instance, we transform given $x_i$ in $A$ to another image from different domains such as $E, C, I$. If we use different images belonging to these domains to transform, we can define new generated input dataset as following list $x^+_{i} = \left[x_{i},x^E_{i},x^C_{i},x^I_{i}\right]$ where $x_{i}$ is the original image and $x^+_{i}$ is a new input list including the original image. 

Transferring sensitive information while keeping prior information of the image is possible via adversarial methods, as they are capable of generating images from the training data distribution. To show that, we propose a solution of sensitive attribute transformation while keeping prior information for face recognition and present a new augmented dataset, $D^+_{image}= \left[x_{i}, x^{A}_{i}, x^{C}_{i}, x^{I}_{i},\dotsc x_{i}, x^{E}_{i}, x^{C}_{i}, x^{I}_{i}, \dotsc, x_{n}, x^{A}_{n}, x^{E}_{n}, x^{C}_{n}, \right]$. In the next Section \ref{subsec:cyclegan} we present our approach to the image synthesise process to obtain $D^+_{image}$.

\subsection{Adversarial Image-to-Image Transfer}
\label{subsec:cyclegan}
Our solution transforms these sensitive attributes using a cyclic adversarial domain transfer approach, CycleGAN \cite{zhu2017unpaired}. We assume that learning a mapping function between two different race groups domain reduces the dependency on sensitive features.

For example, given an African face image $x_i \in A$, and a Caucasian image $x_j \in C$, we assume that the two different data distributions from these image race groups $x_{i} \sim p_{data}(x_{i})$ and $x_{j} \sim p_{data}(x_{j})$ can be transferable between each other. To map these two distributions between domain $A$ and $C$, we introduce two mapping functions $F$ and $G$, respectively from African to Caucasian domains and from Caucasian to African domains using CycleGAN \cite{zhu2017unpaired}. Within a GAN framework, these two directional transformations need two discriminators $D_A$ and $D_C$, to distinguish between $x_{i}$ and $F(x_{j})$, $x_{j}$ and $G(x_{i})$, respectively. Moreover, as an additional control on adversarial training, a cycle-consistency loss is introduced to ensure that the mapping function can transfer an individual input $x_{i}$ to the desired output $x_{j}$.
\begin{align}
\label{eq:adversarialloss}
L_{GAN}(G, D_C, A, C ) & = \mathbb{E}_{x_j} \sim p_d(x_j)\left[log D_C(x_j)\right] \\ \nonumber
 & + \mathbb{E}_{x_i} \sim p_d(x_i)\left[log(1 - D_C(G(x_i))\right]
\end{align}

For the first part of race transformation, an adversarial loss is used as defined in Equation \ref{eq:adversarialloss} where $A$ and $C$ are the African and Caucasian group domains, respectively. While the generator $G$ synthesise images using source domain $A$ to associate to target domain $C$, discriminator $D_C$ distinguishes between the real image and $x_j$ from the synthesised image, $G(x_{i})$. The same process is applied with generator $F$ and discriminator $D_A$ to transform domains from $C$ to $A$. 

The key premise of CycleGAN \cite{zhu2017unpaired} is a controlled mechanism of adversarial training which allows us to synthesise more accurate images from the desired images in the domain. To achieve this, cycle consistency loss is introduced as defined in Equation \ref{eq:cycleconsistency} , where $F(G(x_i))$ is reconstructed $x_i$ from synthesised $G(x_i)$ new image. In this case, generators $F$ and $G$ are able to reconstruct the original images. The $L1$ norm in this loss measures the difference between the original image and reconstructed image as follows:
\begin{align}
\label{eq:cycleconsistency}
L_{cyc}(G, F) & = \mathbb{E}_{x_i} \sim p_d(x_i) \left[ \parallel { F(G(x_i)) - x_{i} \parallel }_1 \right] \\ \nonumber
 & + \mathbb{E}_{x_j} \sim p_d(x_j) \left[ { \parallel G(F(x_j)) - x_j \parallel }_1 \right]
\end{align}
The overall loss function, as defined in Equation \ref{eq:overall_loss}, consists of two adversarial loss within the cycle-consistency loss where $\lambda$ is a term to control the relative importance of the cycle-consistency loss.
\begin{align}
\label{eq:overall_loss}
L(G, F, D_A, D_C ) & = L_{GAN}(G, D_C , A, C ) \\ \nonumber
&+ L_{GAN}(F, D_A, C, A) \\ \nonumber
&+ \lambda L_{cyc}(G, F)]
\end{align}
Subsequently, overall adversarial training of this objective function aims to solve the following equation:

\begin{align}
\label{eq:minmax}
G^*,F^* = \underset{G,F}{\text{argmin}} \underset{D_A,D_C}{\text{max}} L(G, F, D_A, D_C).
\end{align}

In the intermediate step $G(x_i)$ and $F(x_j)$, the generator encodes features of inputs $x_i$ and $x_j$ and then $F(x_j)$ and $G(x_i)$ decodes back to obtain original images again. With reference to this set of transform Equations \ref{eq:adversarialloss}-\ref{eq:minmax}, we can transform both, domain $A$ into domain $C$ and $C$ into $A$ similarly for other domain pairings. 
\subsection{Face Recognition}
\label{subsec:facerecognition}
Recent state-of-the-art face recognition algorithms \cite{liu2017sphereface,wang2018cosface,deng2019arcface,cao2018vggface2} achieve outstanding results for both face verification and identification tasks on public datasets. However, they are not as reliable for real-world racial diversity as their performance is lower for under-represented racial groups \cite{wang2019racial}. 

In Section \ref{subsec:cyclegan}, we presented our proposed approach to address racial bias within face recognition using an adversarial image-to-image transformation technique. To assess this proposed approach, we first present current face recognition loss functions namely Softmax, CosFace, ArcFace that underpin leading state-of-the-art face recognition algorithms \cite{liu2017sphereface,wang2018cosface,deng2019arcface}, then we utilise each of these three methods in conjunction with our cyclic adversarial domain transfer approach. 

The Softmax \cite{liu2017sphereface}, CosFace \cite{wang2018cosface} and ArcFace \cite{deng2019arcface} methods are based on loss functions that operate on the outputs of the last fully connected layer of the selected backbone Deep Convolutional Neural Network \cite{he2016deep} (DCNN). In essence, after feeding an image forward through a DCNN, we obtain the feature space representation of the image. These loss functions enforce different representations of features to predict if they belong to a given subject. First, Softmax loss is formulated as follows, 
\begin{align} \label{eq:softmax}
\mathcal{L}_{1} = - \frac{1}{N} \sum_{i=1}^{N} \log \frac{e^{W_{y_i}^{T} z_i + b_{y_i}}}{\sum_{j=1}^{n}{e^{W_{j}^{T} z_i + b_j}}}
\end{align}
where $z_i$ is the feature representation of the image $x_i\in \mathbb{R}^{d}$ in the dataset $D$ belonging to ${y}_i$-th subject class. The number of samples is $N$ labelled with $n$ classes. $W_j$ is the j-th column of the weights and $b_j$ is the j-th column of the bias term in the last fully-connected layer. Weights and bias term dimensions are $W \in \mathbb{R}^{dxn}$ and $b_j \in \mathbb{R}^n$, respectively. 

Softmax loss \cite{liu2017sphereface} is one of the most widely used objective function to learn optimal feature representations from images. It discriminates deep representations from different classes by maximizing the posterior probability of the ground-truth class. Once large-scale datasets have high similarity on intra-class samples and diversity on inter-class samples, Softmax loss entangles features \cite{he2019softmax}.  To address this problem, CosFace \cite{wang2018cosface} proposes to use both norm and angle of the feature representation to contribute to the posterior probability such that:
\begin{align} \label{eq:cosface}
\mathcal{L}_{2} = - \frac{1}{N} \sum_{i=1}^{N}  \log \frac{e^{{\parallel z\parallel}(\cos(\theta_{y_i,i})-m)}}{e^{{\parallel z\parallel}(\cos(\theta_{y_i,i})-m)} + \sum_{j \neq y_i}^{n}{e^{{\parallel z\parallel}(\cos(\theta_{j,i}))}}}
\end{align}
where $\cos(\theta_j , i)={W^T}_j z_i$ and 
$z_i$, $y_i$,$n$ $N$, $W_i$ denote $i$-th feature  representation with all other definitions as per previously defined. For CosFace loss, the bias term is removed, and the weights $W$ and embeddings $z$ are normalized using the $L_2$ normalization. To cope with incorrect classified samples, a cosine margin $m$ is applied to the classification boundary.

An alternative loss function, ArcFace \cite{deng2019arcface} differs from CosFace \cite{wang2018cosface} based on its distinct margin. ArcFace has more accurate geodesic distance due to it has constant linear angular margin, $m$ penalty throughout the interval while CosFace has a nonlinear angular margin. It also normalizes the weights and embeddings and fixes the bias term to zero. In Equation \ref{eq:arcface}, the ArcFace loss function is formulized as follows:
\begin{align} \label{eq:arcface}
\mathcal{L}_{3} = - \frac{1}{N} \sum_{i=1}^{N} \log \frac{e^{{\parallel z\parallel}(\cos(\theta_{y_i,i}+m))}}{e^{{\parallel z\parallel}(\cos(\theta_{y_i,i}+m))} + \sum_{j \neq y_i}^{n}{e^{{\parallel z\parallel}(\cos(\theta_{j,i}))}}}
\end{align}
where all definitions are as per Equation \ref{eq:cosface}. Overall the key Softmax, CosFace \cite{wang2018cosface} and ArcFace \cite{deng2019arcface} differences lie in their use of deep feature representation, weight vectors and approach to their margin penalty. Within the scope of this study, we only use these methods as experimental vehicles to illustrate our per-subject data augmentation methodology to address face recognition race bias within such state-of-the-art face recognition algorithms.

An overview of our approach is shown in Figure \ref{fig:overview}. Figure \ref{fig:overview} (a) describes the real-world dataset imbalanced distribution for different racial groups. As an initial experimental exercise, we, downsample this imbalanced distribution to understand the relationship between bias and data. In Figure \ref{fig:overview} (b), we explain the image transformation process for one exemplar Asian subject. We introduce our $x_i$ to three different CycleGAN and obtain three different synthesised images $x^A_i,x^C_i,x^I_i$. Subsequently, the training dataset has changed, and then we use our newly augmented dataset for face recognition using algorithms with Softmax, CosFace \cite{wang2018cosface}, ArcFace \cite{deng2019arcface} in Figure \ref{fig:overview} (c). 
    \section{Experimental Setup}
\label{sec:experimental-setup}
This section provides overview of our experimental evaluation in terms of the face recognition datasets used, the race classification used for racial annotation and the implementation details of our proposed approach.
\subsection{Datasets}
\label{sec:datasets}
To validate our approach, we utilise BUPT-Transferface \cite{wang2019racial}, VGGFace2 \cite{cao2018vggface2} and RFW \cite{wang2019racial}.

\vspace{-0.50cm}
\paragraph{BUPT-Transferface} \cite{wang2019racial} provides 50K African, Asian and Indian face images and over 460K Caucasian face images. We use BUPT-Transferface dataset for two different purposes: (i) race transfer, (ii) race classification.
\vspace{-0.50cm}
\paragraph{VGGFace2} \cite{cao2018vggface2} contains 3.3M+ images for over 9K subjects (8631 subjects training examples, 500 testing examples). We train the face recognition methods which we introduced in Section \ref{subsec:facerecognition} on VGGFace2. 
\vspace{-0.50cm}
\paragraph{VGGFace2 1200} is a subsampled version of VGGFace2 which is racially balanced and contains 300 subjects per-race. We evaluate our approach on both VGGFace2 1200 and VGGFace2.

\vspace{-0.50cm}
\paragraph{Racial Faces in-the-Wild (RFW)} \cite{wang2019racial} is a face verification test set which provides 6K pairs of images for each race. We compare the verification accuracy of our proposed approach on different races using the same protocol in \cite{huang2008labeled}.

\subsection{Race Annotations}
\label{sec:raceclassifier}
We obtain racial annotation labels for VGGFace2 \cite{cao2018vggface2} dataset using fine-grained classification to solely support our development of a technique to mitigate bias.

\begin{figure}[ht]
\centering
\includegraphics[width=\columnwidth]{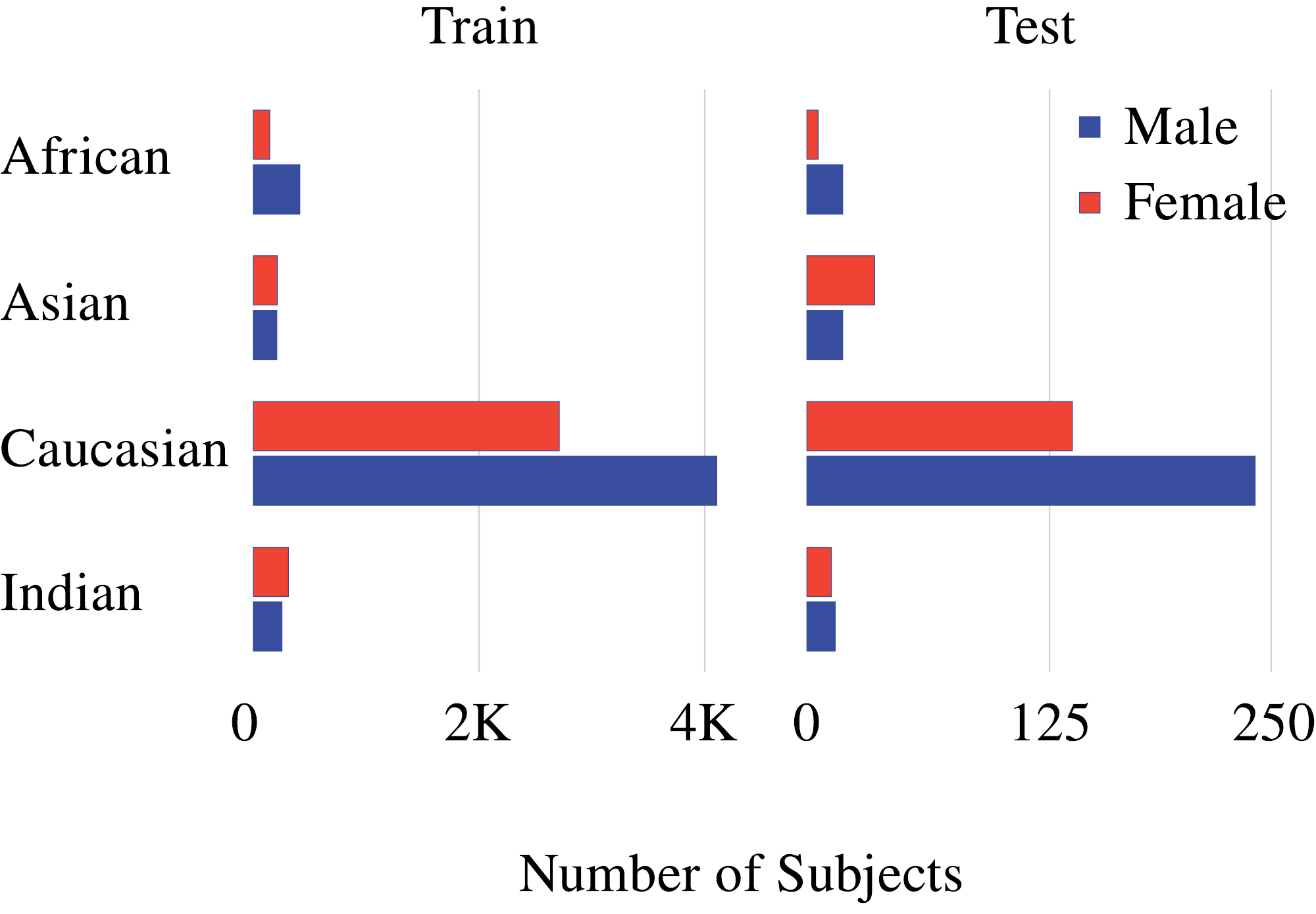}
\caption{\label{fig:vgg_stats} VGGFace2 dataset gender and race distribution for train and test.}
\end{figure}

The work of \cite{hu2019see} proposes attention-guided data augmentation to improve the spatial representation of discriminative image parts using its cropping and dropping mechanism. We adopt this solution for a race classification problem where discriminative image parts are facial attributes of eyes, nose, mouth, and forehead. Via this approach \cite{hu2019see}, we obtain racial annotations of VGGFace2 \cite{cao2018vggface2} and we manually check the least certain subjects according to the majority of image labels for each subject and additionally exclude some subjects who are not in the four-race set $\{Caucasian, African, Asian, Indian\}$. After this semi-automatic process, the subject distribution for training and testing sets is shown in Figure \ref{fig:vgg_stats} whereby the inherent racial and gender imbalance is clearly illustrated.


\subsection{Race Transfer}
Our proposed image-to-image transformation approach creates a new dataset $D^+_{image}$, to transfer race attributes from one race group to another. To achieve that, we define separate mappings for each pair of the four different race groups. The set of 12 mappings are: \{African $\rightarrow$ Asian, African $\rightarrow$ Caucasian, African $\rightarrow$ Indian, Asian $\rightarrow$ African, Asian $\rightarrow$ Caucasian, Asian $\rightarrow$ Indian, Caucasian $\rightarrow$ African, Caucasian $\rightarrow$ Asian, Caucasian $\rightarrow$ Indian, Indian $\rightarrow$ African, Indian $\rightarrow$ Asian, Indian $\rightarrow$ Caucasian\}. As our CycleGAN based approach provides two-way transformations between source and target domains, we train six models to find these two directional mappings following the approach outlined in Section \ref{subsec:cyclegan}.

For training, we generate 25K image pairs using the BUPT-Transfer \cite{wang2019racial} dataset. All face images are aligned and have a size of $256 \times 256$. To avoid gender domain differences, we only match images of the same gender as pairs.
Using these six CycleGAN models, we synthesise new images and denote extended dataset as VGGFace2 1200 Races \cite{cao2018vggface2} which contains the original VGGFace2 1200 images and synthesised race images. Each image has three different transformed images that belong to other race domains in addition to the original. As a result, we partially absorb the downsampling effect on VGGFace2 1200. Subsequently, we synthesise all non-Caucasians images on original VGGFace2 and call the new dataset VGGFace2 8631 Races, $D^+_{image}$. We do not transform Caucasian images to other racial domains; they are already dominant in the original dataset.

\subsection{Face Recognition}

We train Softmax \cite{liu2017sphereface}, CosFace \cite{wang2018cosface} and ArcFace \cite{deng2019arcface} loss over a common convolutional neural network backbone, Resnet \cite{he2016deep} using proposed augmented datasets; VGGFace 2 1200, VGGFace 2 8631. We utilise Resnet100 from \cite{deng2019arcface} with ${BatchNorm-Dropout-FC-BatchNorm}$ structure to get the final 512-D feature space representation after the last convolutional layer. For ArcFace margin on VGGFace2 we set $m$ to $0.3$.
    \section{Results}
\label{sec:results}
\begin{figure*}[ht]
\begin{center}
\includegraphics[width=\textwidth]{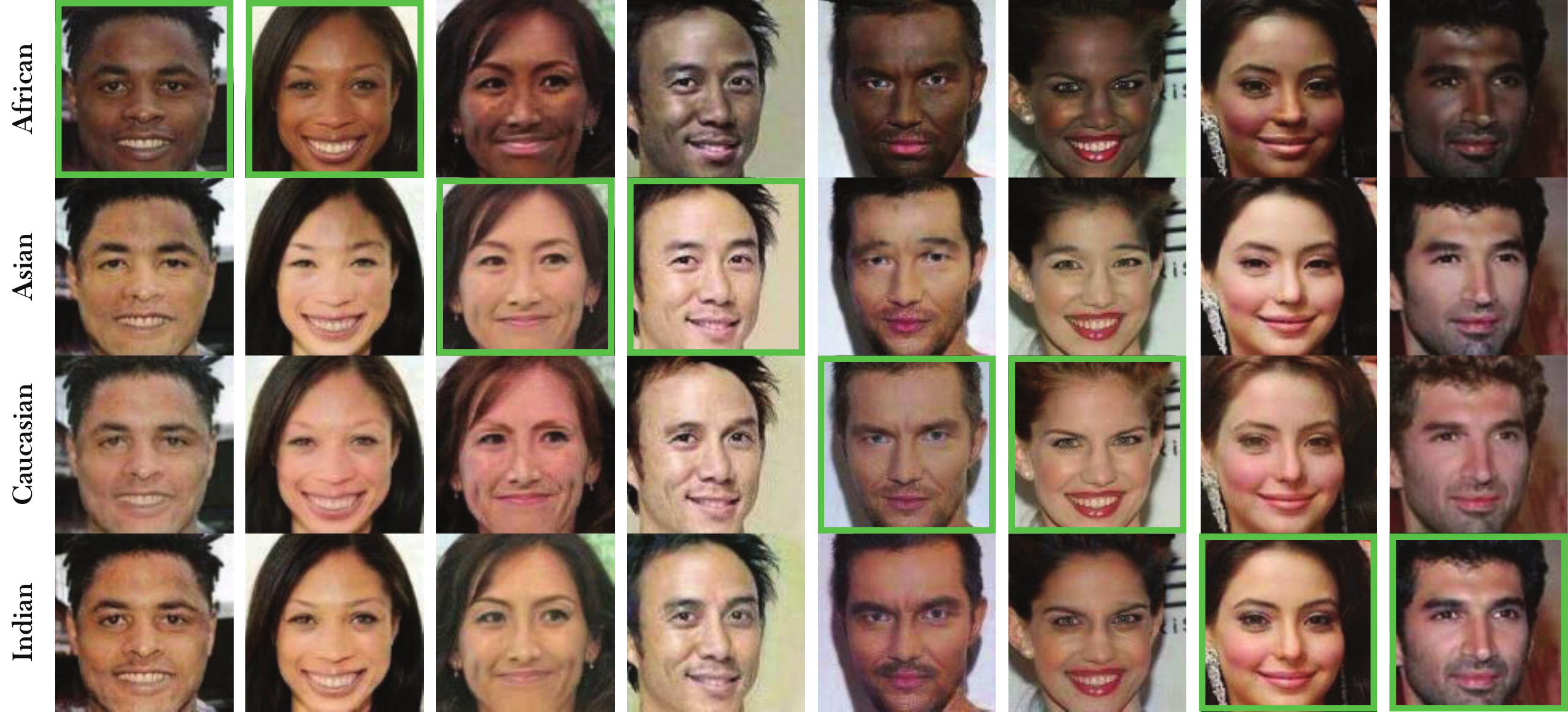}
\end{center}
\vspace{-0.5cm}
\begin{center}
\includegraphics[width=\textwidth]{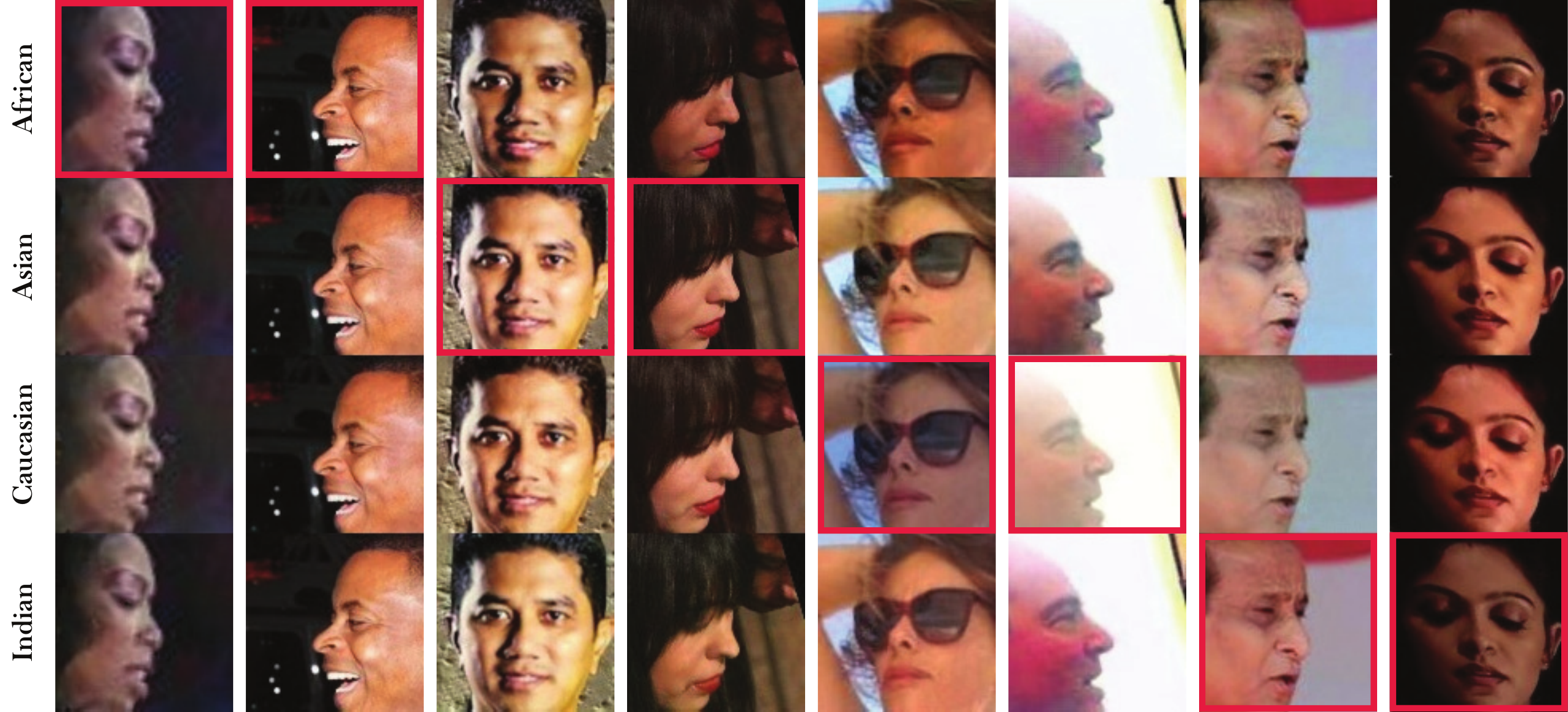}
\end{center}
\caption{A selection of successful (top) and failure (bottom) examples of the CycleGAN racial domain transformation of VGGFace2 dataset. Each column contains an original and sythesised face images of the same subject where the green (top) and red (bottom) borders indicate the original image and the corresponding race labels are laid out on the y-axis.} 
\label{fig:cycleganbestworst}
\end{figure*}

To evaluate the performance of the proposed approach, we use LFW face verification protocol \cite{huang2008labeled}, which measures whether two images belong to the same subject or not.

We assess synthesised image quality by feeding them through a race classifier introduced in Section \ref{sec:raceclassifier}. We show examples of the correctly classified images and the misclassified images in Figure \ref{fig:cycleganbestworst} (top and bottom parts are separated). Each column of Figure \ref{fig:cycleganbestworst} show an image transformation example where the original image is represented with green and red borders, and synthesised images are laid in the corresponding racial domain label in the y-axis. As can be seen in the bottom part of Figure \ref{fig:cycleganbestworst}, image transformation is prone to fail on poor illumination and pose variations.

\begin{table}[!ht]
\begin{center}
\resizebox{\columnwidth}{!}{%

\begin{tabular}{lllcccccc}
\hline
\multirow{2}{*}{Loss} & \multirow{2}{*}{Training Dataset} & \multirow{2}{*}{LFW} & \multicolumn{6}{c}{RFW}                                                                            \\
                      &                                   &                      & African        & Asian          & Caucasian      & Indian         & AVG            & STDV          \\ \hline
Softmax               & VGGFace2 1200                     & 96.13                & 69.10          & 73.70          & 79.25          & 76.78          & 74.71          & 4.37          \\
Softmax               & VGGFace2 1200 Races               & \textbf{96.27}       & \textbf{70.65} & \textbf{75.68} & \textbf{80.27} & \textbf{78.28} & \textbf{76.22} & \textbf{4.16} \\ \hline
CosFace               & VGGFace2 1200                     & 98.16                & 82.78          & 82.68          & 87.53          & 85.41          & 84.60          & 2.33          \\
CosFace               & VGGFace2 1200 Races               & \textbf{98.65}       & \textbf{83.22} & 83.23          & \textbf{87.95} & \textbf{85.77} & \textbf{85.04} & \textbf{2.28} \\ \hline
Arcface               & VGGFace2 1200                     & 98.16                & 80.91          & 81.78          & \textbf{86.86} & 83.70          & 83.31          & 2.64          \\
Arcface               & VGGFace2 1200 Races               & \textbf{98.63}       & \textbf{81.28} & 82.83          & 85.95          & \textbf{84.72} & \textbf{83.69} & \textbf{2.06} \\ \hline
\end{tabular}
}
\end{center}
\caption{Verification performance (\%) of Softmax, CosFace, and ArcFace with ResNet-101 \cite{he2016deep} on LFW \cite{huang2008labeled} and RFW \cite{wang2019racial} when trained on VGGFace2 1200 and proposed VGGFace2 1200 Races datasets.}
\label{tab:vgg_balanced}
\end{table}

For face recognition, we first test our performance on \textbf{balanced datasets} VGGFace2 1200 and VGGFace2 1200 Races. We compare our results on RFW \cite{wang2019racial} using three different loss functions; Softmax, CosFace  \cite{wang2018cosface} and ArcFace \cite{deng2019arcface} as shown in Table \ref{tab:vgg_balanced}. Proposed facial image augmentation approach improves performance in all three methods by 0.38-1.51 \%. As non-Caucasian results are improved, the standard deviation among groups is decreased. We also share LFW results in Table \ref{tab:vgg_balanced} to show the improvement of our solution on the imbalanced dataset. Second, we use the \textbf{imbalanced dataset} with the ArcFace as shown in Table \ref{tab:vgg_unbalanced}. While LFW verification performance remains the same, RFW African and Asian performances are improved, and the standard deviation declines from 2.91 to 2.45.

\begin{table}[ht]
\begin{center}
\resizebox{\columnwidth}{!}{%
\begin{tabular}{llllllll}
\hline
\multirow{2}{*}{Training Dataset} & \multicolumn{1}{c}{\multirow{2}{*}{LFW}} & \multicolumn{6}{c}{RFW}                                                                            \\
                                  & \multicolumn{1}{c}{}                     & African        & Asian          & Caucasion      & Indian         & Average        & STDV          \\ \hline
VGGFace2                        & \textbf{99.51}                           & 89.45          & 87.61          & \textbf{94.71} & \textbf{91.21} & \textbf{90.75} & 2.91          \\
VGGFace2 8631 Races                   & \textbf{99.51}                           & \textbf{90.10} & \textbf{87.73} & 93.72          & 90.50          & 90.51          & \textbf{2.45} \\ \hline
\end{tabular}}
\end{center}
\caption{Verification performance (\%) of ArcFace using ResNet-101 \cite{he2016deep} trained on VGGFace2 \cite{cao2018vggface2} and VGGFace2 8631 Races with syntesised images of non-Caucasian subjects on VGGFace2, tested on LFW \cite{huang2008labeled} and RFW \cite{wang2019racial}.}
\label{tab:vgg_unbalanced}
\end{table}

\subsection{Ablation Study}
\noindent
{\bf Q:} This study provides experiments on both balanced and imbalanced training datasets. Why do you not use only the imbalanced datasets? Does balancing datasets help to decrease bias?  \\
{\bf A:} Imbalanced data may seem to be the main reason for face recognition bias. However, when we train algorithms on completely equally distributed data, the results still appear to exhibit performance bias. To show this, we downsample VGGFace2 and obtain 1000 subjects with 100 images on each subject. We also keep the race and gender groups balanced. As shown in Table \ref{tab:rfw_verify_balanced}, there is still about eight per cent gap between African and Caucasian on average. Another study experiments on a large and nearly balanced dataset and again differs on Caucasians and non-Caucasians \cite{wang2019skewness}. Subsequently, we focus on a  novel per-subject racial data balancing approach to understanding its impact on the face recognition bias.   \medskip\\
\noindent
\noindent
{\bf Q:} How does the training of CycleGAN affect overall accuracy? \\
{\bf A:} We assess the quality of our synthesised images by testing them using a race classifier (Section \ref{sec:raceclassifier}). We would expect the race classifier to recognise them as the correct transformed racial label. Our overall accuracy is 49\% across all transformations, but when we increase this accuracy using more pairs, and longer training, this results in an overall reduction in face recognition performance.
The trade-off is complex because after transforming the main racial attributes of the face such as skin colour, eye structure and hair colour, CycleGAN proceeds to translate all facial features including those which implicitly encode unique subject identity. Other notable negatives are variations in pose and illumination on the synthesised images which could alternatively be addressed via \cite{karras2019style} in future work. 

\begin{table}[tbp]
\begin{center}
\resizebox{\columnwidth}{!}{%
\begin{tabular}{@{}lllllll@{}}
\toprule
Method  & African & Asian & Caucasian & Indian & AVG   & STDV \\ \midrule
Softmax  & 67.95   & 73.5  & 77.77     & 75.78  & 73.75 & 4.24 \\
CosFace & 77.15   & 78    & 82.8      & 80.42  & 79.59 & 2.55 \\
ArcFace & 74.75   & 77.63 & 83.18     & 80.97  & 79.13 & 3.71 \\ \bottomrule

\end{tabular}
}
\end{center}

\caption{RFW dataset verification performance using the LFW protocol \cite{huang2008labeled} for state-of-the-art algorithms trained on per-subject, per-race and per-gender balanced data.}
\label{tab:rfw_verify_balanced}
\end{table}

\section{Conclusion}
\label{sec:conclusion}

Although the usage of face recognition applications is increasing every day, state-of-the-art-methods are still suffering from racial bias in terms of performance. To address this issue, in this study, we explore racial bias in face recognition and present a novel adversarial derived data augmentation methodology. Transferring racial attributes of a human face whilst preserving identity features in the face recognition datasets makes face recognition algorithms more robust and less race-dependant. We demonstrate that our proposed technique improves face recognition accuracy on minority groups by 1\% using imbalanced and balanced training datasets. On our manually balanced dataset, we also compare three significant face recognition variants: Softmax \cite{liu2017sphereface}, CosFace \cite{wang2018cosface} and ArcFace \cite {deng2019arcface} loss functions with a common convolutional neural network backbone ResNet-101 \cite{he2016deep}. Although illumination, pose, and light challenge the quality of the image transformation; our technique not only improves the overall face recognition accuracy but also suppresses inter-group performance variation.

{\small
\bibliographystyle{lib/ieee_fullname}
\bibliography{ref/egbib}
}

\end{document}